\documentclass{article} 
\usepackage{iclr2027_conference,times}

\usepackage{amsmath,amsfonts,bm}

\def\eqref#1{equation~\ref{#1}}

\def\1{\bm{1}}

\DeclareMathAlphabet{\mathsfit}{\encodingdefault}{\sfdefault}{m}{sl}
\SetMathAlphabet{\mathsfit}{bold}{\encodingdefault}{\sfdefault}{bx}{n}

\usepackage{graphicx}
\usepackage{booktabs}
\usepackage{wrapfig}
\usepackage{tikz}
\usetikzlibrary{arrows.meta,positioning}
\usepackage{hyperref}
\usepackage{url}

\title{Random Recursive Models}

\author{
Jama Hussein Mohamud$^{1,2}$ \quad Mirco Ravanelli$^{1,2,3}$\\
{\small $^1$Mila -- Quebec AI Institute \quad
$^2$Universit\'e de Montr\'eal \quad $^3$Concordia University}
}

\iclrfinalcopy 
\begin{document}

\maketitle
\lhead{}

\begin{abstract}
Recursive models create computational depth through parameter reuse, offering
a parameter-efficient alternative to increasing model size. However, most
recursive models repeatedly apply one learned transformation or a prescribed
sequence of transformations, restricting computation to a fixed layer order.
We introduce the Random Recursive Model (RRM), which maintains a pool of $L$
learned layers and performs $T$ recursive steps by sampling one layer
independently with replacement for each example and step. This enables flexible
layer reuse while retaining the parameter efficiency of recurrence. We evaluate
RRM on challenging reasoning tasks, where it matches or exceeds the baselines,
often with 50--75\% fewer parameters. RRM can vary its
depth at inference, including beyond that seen during training, without
retraining or adding parameters, improving tasks that benefit from deeper
iterative computation. RRM also supports Monte Carlo inference and
probabilistic test-time scaling, both of which improve performance without
retraining. These insights may open new directions in neural network
architecture design.

\end{abstract}

\section{Introduction}

Recursive models have recently obtained strong results on challenging
reasoning tasks \citep{geiping2025scalingtesttimecomputelatent,wang2025hierarchicalreasoningmodel,
jolicoeurmartineau2025morerecursivereasoningtiny}. By reusing parameters across depth, they can
allocate many computational steps to an example without requiring a distinct
learned layer at every step. This makes recursion a direct way to separate
model size from computation, including computation allocated through test-time
scaling \citep{snell2024scalingllmtesttimecompute,zhu2026scalinglatentreasoninglooped}.

Most recursive models nevertheless use a fixed-order recurrence. They either
repeat one learned transformation or repeatedly apply a set of transformations
in a prescribed order. Recent works add further structure
through fixed update schedules, separate high- and low-level modules, distinct
answer and reasoning states, or provisional-answer feedback
\citep{wang2025hierarchicalreasoningmodel,jolicoeurmartineau2025morerecursivereasoningtiny}. We study a different
and simpler form of recursion, which we call the \emph{Random Recursive Model}
(RRM), in which learned layers form a pool of reusable transformations and any
layer can be reused at any point in the computation.

An RRM contains $L$ learned layer transformations and randomly reuses them
over $T$ applications, or recursive steps. Unlike standard recurrence, which creates depth by applying
the same transformation or the same fixed sequence, RRM can revisit the same
layer or apply any of the other learned layers at each application. During
both training and inference, at each of the $T$ recursive steps, one layer is
sampled uniformly and independently with replacement for each example.
RRM therefore retains the parameter sharing of recurrence while
allowing up to $L^T$ possible layer sequences, without imposing a fixed
recurrent order. The number of learned layers $L$ determines model size, while $T$
determines computational depth and can be increased or decreased at test time. 

\begin{figure}[t]
    \centering
    \definecolor{rrmblue}{HTML}{4C78A8}
    \definecolor{rrmorange}{HTML}{F28E2B}
    \definecolor{rrmgreen}{HTML}{59A14F}
    \definecolor{rrmpurple}{HTML}{B279A2}
    \begin{tikzpicture}[
        x=0.93cm,
        y=0.82cm,
        >=Latex,
        state/.style={font=\small},
        block/.style={
            draw,
            rounded corners=1.5pt,
            minimum width=0.72cm,
            minimum height=0.52cm,
            inner sep=1.5pt,
            font=\small\bfseries,
            text=white
        },
        flow/.style={->, semithick, draw=black!65},
        rowlabel/.style={font=\small\bfseries, anchor=east, align=right},
        drawlabel/.style={font=\scriptsize, text=black!60, anchor=east}
    ]
        \node[rowlabel] at (0,2.35) {Fixed-order depth};
        \node[state] (ff-in) at (0.75,2.35) {$h_0$};
        \node[block,fill=rrmblue]   (ff-1) at (2.00,2.35) {$f_1$};
        \node[block,fill=rrmorange] (ff-2) at (3.35,2.35) {$f_2$};
        \node[block,fill=rrmgreen]  (ff-3) at (4.70,2.35) {$f_3$};
        \node[block,fill=rrmpurple] (ff-4) at (6.05,2.35) {$f_4$};
        \node[state] (ff-out) at (7.28,2.35) {$h_4$};
        \draw[flow] (ff-in) -- (ff-1);
        \draw[flow] (ff-1) -- (ff-2);
        \draw[flow] (ff-2) -- (ff-3);
        \draw[flow] (ff-3) -- (ff-4);
        \draw[flow] (ff-4) -- (ff-out);

        \node[rowlabel] at (0,1.25) {Fixed-recursive depth};
        \node[state] (rec-in) at (0.75,1.25) {$h_0$};
        \node[block,fill=rrmblue] (rec-1) at (2.00,1.25) {$f$};
        \node[block,fill=rrmblue] (rec-2) at (3.35,1.25) {$f$};
        \node[block,fill=rrmblue] (rec-3) at (4.70,1.25) {$f$};
        \node[block,fill=rrmblue] (rec-4) at (6.05,1.25) {$f$};
        \node[state] (rec-out) at (7.28,1.25) {$h_4$};
        \draw[flow] (rec-in) -- (rec-1);
        \draw[flow] (rec-1) -- (rec-2);
        \draw[flow] (rec-2) -- (rec-3);
        \draw[flow] (rec-3) -- (rec-4);
        \draw[flow] (rec-4) -- (rec-out);

        \node[rowlabel] at (0,0.15) {RRM};
        \node[drawlabel] at (0,-0.32) {$\mathbf r=(2,1,2,4)$};
        \node[state] (rrm-in) at (0.75,0.15) {$h_0$};
        \node[block,fill=rrmorange] (rrm-1) at (2.00,0.15) {$f_2$};
        \node[block,fill=rrmblue]   (rrm-2) at (3.35,0.15) {$f_1$};
        \node[block,fill=rrmorange] (rrm-3) at (4.70,0.15) {$f_2$};
        \node[block,fill=rrmpurple] (rrm-4) at (6.05,0.15) {$f_4$};
        \node[state] (rrm-out) at (7.28,0.15) {$h_4$};
        \draw[flow] (rrm-in) -- (rrm-1);
        \draw[flow] (rrm-1) -- (rrm-2);
        \draw[flow] (rrm-2) -- (rrm-3);
        \draw[flow] (rrm-3) -- (rrm-4);
        \draw[flow] (rrm-4) -- (rrm-out);

        \node[font=\scriptsize,text=black!60,anchor=west] at (1.55,-0.58)
            {sampled with replacement from $\mathcal F=\{f_1,f_2,f_3,f_4\}$};
    \end{tikzpicture}
    \caption{\textbf{Fixed-order depth, fixed-recursive depth, and RRM.}
    Four applications under the three depth constructions. Fixed-order depth
    applies each layer once in its prescribed position. Fixed-recursive depth
    repeatedly applies a prescribed sequence of learned functions. Here, the
    repeated sequence contains one function ($L=1$); more generally, it may
    contain multiple layers and be surrounded by fixed-order layers. RRM samples
    from the complete set of learned layers at every step, allowing both
    reordering and reuse. Colors identify shared parameters.}
    \label{fig:rrm-computation}
\end{figure}

Across our experiments, we start from established fixed-order baselines and
replace their backbones with RRM while retaining the task formulation and core
model components. Across all five tasks, the strongest RRM configuration
matches or exceeds the corresponding baseline. In controlled image
classification and autoregressive logic puzzles, RRM does so with roughly
50--75\% fewer parameters. The same RRM construction also matches or exceeds
existing methods on more difficult reasoning tasks without any structured, 
hierarchical or prescribed structure.

We further study how RRM scales with the number of learned layers $L$ and the
number of applications $T$. RRM therefore naturally supports both training-time
scaling and test-time scaling, since $T$ can be increased without adding
learned parameters. We find that additional applications improve
performance when a task benefits from longer iterative computation, while
gains plateau when fewer applications suffice and performance can degrade when
test-time depth extends far beyond training depth. RRM is also compatible
with Monte Carlo inference and probabilistic test-time scaling, both of which
improve performance without retraining. With the latter, RRM achieves the best
result among the three evaluated models.

\section{Background}
\label{sec:background}

Most deep networks execute the learned transformations in a fixed order. This is
true both for conventional feed-forward networks, which assign different
parameters to different depths, and for modern recursive networks, which reuse
parameters across depth. The two families differ in parameter sharing, but
both prescribe the sequence of transformations before seeing an example.

\subsection{Fixed-order depth}

A feed-forward network of depth $L$ maps an encoded input $h_0=E(x)$ through
an ordered sequence of functions,
\begin{equation}
    h_{\ell+1}=f_{\ell+1}(h_\ell),
    \qquad \ell=0,\ldots,L-1.
    \label{eq:dense-computation}
\end{equation}
The computation is therefore the fixed composition
$f_L\circ\cdots\circ f_1$. Every example applies each function once and in the
same order. Transformers are a prominent instance of this design. Each layer
has its own parameters and occupies a predetermined position in the stack
\citep{vaswani2023attentionneed}. Increasing depth ordinarily adds both computation
and parameters.

\subsection{Fixed-recursive depth}

Recursive models instead create depth by applying a learned transformation
repeatedly. In its simplest form,
\begin{equation}
    h_{t+1}=f(h_t,E(x)),
    \qquad t=0,\ldots,T-1,
    \label{eq:fixed-recurrence}
\end{equation}
where the encoded input may optionally be injected at each step. Parameter
sharing decouples computational depth $T$ from model size. This principle
underlies the Universal Transformer \citep{dehghani2019universaltransformers}, looped
Transformers \citep{yang2024loopedtransformersbetterlearning}, and recurrent-depth language models
\citep{geiping2025scalingtesttimecomputelatent}. Recent reasoning systems similarly obtain large
effective depth from small recurrent modules, including the Hierarchical
Reasoning Model (HRM) \citep{wang2025hierarchicalreasoningmodel} and the Tiny
Recursive Model (TRM)
\citep{jolicoeurmartineau2025morerecursivereasoningtiny}.

The recurrence becomes explicit when we unroll the computation. Define
$f_x(h)=f(h,E(x))$ for a fixed input $x$. After $T$ steps,
\begin{equation}
    h_T
    = \underbrace{f_x\circ f_x\circ\cdots\circ f_x}_{T\text{ times}}(h_0)
    = f_x^{\circ T}(h_0).
    \label{eq:unrolled-recurrence}
\end{equation}
For example, four recursive steps apply
$f_x\circ f_x\circ f_x\circ f_x$, whereas a four-layer feed-forward network
applies $f_4\circ f_3\circ f_2\circ f_1$. The recursive model therefore follows a
prescribed order, repeatedly applying the same learned function or fixed
sequence of functions.

\section{Random Recursive Models}
\label{sec:method}

We introduce the \emph{Random Recursive Model} (RRM). RRM retains the input
encoder $E$, all $L$ learned layers of the backbone
$\mathcal{F}=\{f_1,\ldots,f_L\}$, and the task-specific output head $H$.
Unlike fixed-order feed-forward depth and fixed-recursive depth, RRM makes
every backbone layer reusable rather than applying all layers once in their
prescribed order or repeatedly following a fixed sequence. Given an input $x$, the encoder
initializes the hidden state as
$h_0=E(x)$, although $h_0$ may instead be any task-specific starting state. For
a fixed computational depth $T$, at each step $t=0,\ldots,T-1$, RRM samples
one of the $L$ layers uniformly at random and applies it to the current state.
\begin{equation}
    h_{t+1}=f_{r_t}(h_t,E(x)),
    \qquad r_t\sim\operatorname{Uniform}\{1,\ldots,L\},
    \label{eq:rrm-update}
\end{equation}
where $r_t$ is the layer index sampled at step $t$. Sampling is performed
independently for every example and every recurrent step. The indices are
sampled with replacement, so a function may be applied
multiple times or not at all in a particular computation. The second argument
$E(x)$ denotes optional input reinjection. When the backbone does not inject
the encoded input at every step, Equation~\ref{eq:rrm-update} reduces to
$h_{t+1}=f_{r_t}(h_t)$. After $T$ applications, the task-specific head returns
$\hat{y}=H(h_T)$. RRM does not prescribe a particular layer type. Its layers
may be Transformer blocks or other learned state transformations.

The complete computation for an example is determined by the random sequence
$\mathbf{r}=(r_0,\ldots,r_{T-1})$ and can be written as
\begin{equation}
    h_T
    =
    f_{r_{T-1}} \circ \cdots \circ f_{r_1} \circ f_{r_0}(h_0;E(x)).
    \label{eq:rrm-composition}
\end{equation}

Figure~\ref{fig:rrm-computation} compares RRM with fixed-order depth and
fixed-recursive depth. These architectures differ in how their layers are
ordered and reused. Setting $L=1$ repeats a single learned transformation.
Replacing random layer selection with a prescribed repeating sequence gives
fixed-recursive depth, whether that sequence contains one layer, a subset, or
all $L$ layers.
Setting $T=L$ and $r_t=t+1$ gives fixed-order depth. Fixed-recursive and
fixed-order models can therefore be viewed as special cases of RRM.

RRM retains multiple learned layers while composing them recursively. This
preserves the parameter-efficient depth and adjustable computation of recurrence
while giving each step access to multiple learned state transformations. At
recursive depth $T$, RRM admits $L^T$
possible layer sequences instead of committing to one fixed composition,
providing diverse computational trajectories without an additional learned
mechanism.

\subsection{Training}

Let $p(\mathbf{r})$ denote the uniform distribution over length-$T$ index
sequences. For a dataset
$\mathcal{D}$ and task loss $\ell$, RRM minimizes
\begin{equation}
    \mathcal{L}(\theta)
    =
    \mathbb{E}_{(x,y)\sim\mathcal{D}}
    \mathbb{E}_{\mathbf{r}\sim p(\mathbf{r})}
    \left[
        \ell\!\left(H_\theta(h_T(x;\mathbf{r})),y\right)
    \right].
    \label{eq:rrm-objective}
\end{equation}
We estimate this objective with one independently sampled sequence per example
in each forward pass. Unless stated otherwise, standard backpropagation is
applied through the sampled composition; at a given step, gradients update
only the function that was applied. There is no auxiliary objective or
additional learned mechanism.\footnote{We provide details on learned policies in Appendix~\ref{sec:learned-ordering-appendix}.} Inference uniformly samples a sequence of layers for each sample, similar to training.

\subsection{Training- and Test-Time Scaling}

RRM separates the number of physical backbone layers, $L$, from the number of
recursive applications, $T$. The former determines the parameterized depth of
the backbone, while the latter determines its effective computational depth.
At training time, increasing $T_{\mathrm{train}}$ unrolls a longer random composition and
allocates more computation to each example without introducing new layer
parameters. Sampling with replacement makes reuse possible whenever $T>1$ and
guarantees that at least one layer is reused when $T>L$.

The recursive depth may also be changed after training. Given a model trained
with depth $T_{\mathrm{train}}$, we can evaluate the same parameters using a
different depth $T_{\mathrm{test}}$. Increasing $T_{\mathrm{test}}$ allocates
more inference computation while the model size remains fixed. Allocating
additional computation at inference is a widely studied approach to scaling
reasoning \citep{snell2024scalingllmtesttimecompute,geiping2025scalingtesttimecomputelatent,zhu2026scalinglatentreasoninglooped}.
Thus, RRM supports compute scaling during both training and inference by varying
$T_{\mathrm{train}}$ or $T_{\mathrm{test}}$ without changing model size.
RRM also supports Monte Carlo inference and probabilistic test-time scaling,
which are both described in Appendix~\ref{sec:inference-procedures-appendix}.

\section{Experimental Setup}
\label{sec:experiments}

We evaluate RRM in three settings of increasing reasoning difficulty. For each
task, we follow the task formulation, data splits, model components,
training protocol, and evaluation procedure of established baselines. Our main
change is to replace the fixed-order backbone with an RRM backbone. Any
further task-specific changes are stated below. In every experiment, one layer index
is sampled independently for each example and each application. Unless noted
otherwise, the number of applications is fixed within a run. Where separate
validation and test splits are available, we use validation performance for
checkpoint selection and evaluate the selected model on the disjoint test set.
Unless stated otherwise, RRM results are reported as mean $\pm$ sample
standard deviation over three evaluation seeds for a fixed checkpoint.

\paragraph{CIFAR-10.}
We use CIFAR-10 as a controlled toy experiment \citep{krizhevsky2009learning}. The
baseline is a custom eight-layer fixed-order Vision Transformer following
\citet{dosovitskiy2021image}, adapted to CIFAR-10 and trained under the same
protocol as RRM. RRM replaces its eight-layer backbone with two reusable
Transformer layers ($L=2$). For training-time scaling, we train models with
$T_{\mathrm{train}}\in\{4,8,12,16\}$. For test-time scaling, we fix
models trained with $T_{\mathrm{train}}\in\{8,16\}$ and evaluate
$T_{\mathrm{test}}\in\{4,8,12,16,20,24\}$. We additionally evaluate Monte
Carlo inference with $M=16$ for every combination of
$T_{\mathrm{train}}\in\{4,8,12,16\}$ and
$T_{\mathrm{test}}\in\{4,8,12,16,20,24\}$. We
separately train an $L=1$,
$T_{\mathrm{train}}=16$ model to study single-layer reuse. Finally, we examine how the class-token
representations of the $T_{\mathrm{train}}=16$ model evolve over
$T_{\mathrm{test}}\in\{8,12,16,20,24\}$ using one sampled sequence per image. We quantify class
separation with a $k$-nearest-neighbor ($k$-NN) classifier using $k=20$ and
visualize the representations with t-SNE~\citep{vandermaaten08a}. Performance
is measured by top-1 accuracy on the official test set. Architecture and
training details are provided in Appendix~\ref{sec:experimental-details-appendix}.

\paragraph{Sudoku and Zebra.}
We train on the processed Sudoku and Zebra datasets publicly released by
\citet{shah2024causallanguagemodelingelicit}.\footnote{\url{https://github.com/kulinshah98/llm-reasoning-logic-puzzles}}
We use their strongest solver-ordered formulation. The causal Transformer has
eight layers and provides the fixed-order baseline. RRM changes only the
backbone composition. On Sudoku, we evaluate $L\in\{2,4\}$ reusable
layers, while Zebra uses $L=2$. Every RRM is trained with
$T_{\mathrm{train}}=16$ applications.
All comparisons use the same 320,000-step training budget. We train on the full
released training sets and evaluate on task-specific held-out splits using
their greedy autoregressive protocol. For Zebra, we additionally evaluate
test-time scaling for the trained RRM with $L=2$ and
$T_{\mathrm{train}}=16$, using $T_{\mathrm{test}}\in\{4,8,12,16,20\}$.

\paragraph{Extreme reasoning tasks.}
For Sudoku-Extreme, Maze-Hard, and ARC-AGI-1
\citep{chollet2019measure}, we follow the experimental setups used by HRM
\citep{wang2025hierarchicalreasoningmodel} and TRM
\citep{jolicoeurmartineau2025morerecursivereasoningtiny}, including the released
datasets, layer components, objective, and optimization settings.
As in HRM and TRM, the fixed puzzle embedding $E(x)$ is supplied at every
recurrent application. Maze-Hard and Sudoku-Extreme use exact-grid accuracy;
ARC-AGI-1 uses Pass@2. We use these tasks to test whether useful recursive
computation can emerge from simple random layer reuse, without the prescribed
recurrent structures of HRM and TRM.

\paragraph{Maze-Hard.}
On Maze-Hard, we use basic RRM with two non-causal Transformer layers
($L=2$). We train models with $T_{\mathrm{train}}\in\{12,16,20\}$ applications
and evaluate them with
$T_{\mathrm{test}}\in\{8,12,16,20,24,28\}$. For the
$T_{\mathrm{train}}=16$ model, we additionally evaluate $T_{\mathrm{test}}=4$.

\paragraph{Sudoku-Extreme and ARC-AGI-1.}
For Sudoku-Extreme and ARC-AGI-1, we augment RRM with deep supervision, as
detailed in Appendix~\ref{sec:deep-supervision-appendix}.
For Sudoku-Extreme, we evaluate $L\in\{1,2\}$ and use
$N_{\mathrm{sup}}=32$ steps with $T_{\mathrm{train}}=21$
applications per step. For test-time scaling, we vary $T_{\mathrm{test}}$ over
$\{11,16,21,26,32\}$.
We also evaluate TRM+RRM, a variant that keeps the entire TRM setup but
replaces only the fixed order of its two layers with RRM's random reuse. For
ARC-AGI-1, we use $L=2$ and vary the number of applications during training and
inference. Further details are provided in
Appendix~\ref{sec:experimental-details-appendix}.

\section{Results}
\label{sec:results}

\subsection{Controlled Toy Image Classification}

\begin{figure*}[t]
  \centering
  \includegraphics[width=\textwidth]{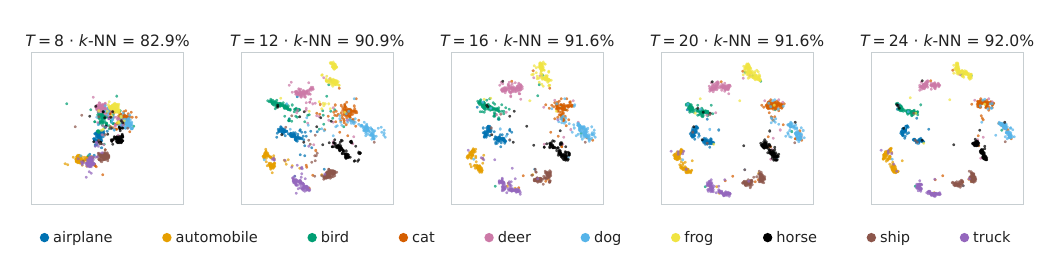}
  \caption{CIFAR-10 representations across recursive depth. We take the
  class-token representations from the RRM trained with $L=2$ and
  $T_{\mathrm{train}}=16$ and plot them at
  $T_{\mathrm{test}}\in\{8,12,16,20,24\}$, using one sampled layer sequence
  per image. Each panel is a t-SNE projection of the same test images, colored by
  class. Both the $k$-NN classification accuracy and the t-SNE projections 
  show that the representations become more class-specific as the number of
  recursive applications increases.}
  \label{fig:cifar10-representations}
\end{figure*}

\begin{wraptable}{r}{0.56\textwidth}
  \vspace{-1.9em}
  \centering
  \caption{CIFAR-10 top-1 accuracy. $M$ denotes Monte Carlo inference samples.}
  \label{tab:cifar10-results}
  \small
  \begin{tabular*}{\linewidth}{@{\extracolsep{\fill}}lccccc@{}}
    \toprule
    Model & $L$ & $T_{\mathrm{train}}$ & Params. & $M$ & Accuracy (\%) \\
    \midrule
    Dense & 8 & 8 & 3.20M & 1 & $92.520 \pm 0.000$ \\
    RRM & 2 & 8  & 0.824M & 1  & $91.403 \pm 0.078$ \\
    RRM & 2 & 16 & 0.824M & 1  & $91.670 \pm 0.155$ \\
    RRM & 2 & 16 & 0.824M & 16 & $92.337 \pm 0.049$ \\
    RRM & 2 & 8  & 0.824M & 16 & $\mathbf{92.597 \pm 0.110}$ \\
    \bottomrule
  \end{tabular*}
  \vspace{-0.6em}
\end{wraptable}

Table~\ref{tab:cifar10-results} compares RRM with the eight-layer fixed-order
dense baseline. With basic inference, the two-layer RRM remains close to the dense
baseline while using roughly one quarter of its parameters. Monte Carlo
inference with $M=16$ closes the gap, and the model trained at
$T_{\mathrm{train}}=8$ slightly surpasses the baseline. Increasing
$T_{\mathrm{train}}$ to 16 provides no consistent benefit, suggesting that
CIFAR-10 saturates at modest computational depth.

Figure~\ref{fig:cifar10-representations} shows that the representations
continue to become more class-specific as we increase recursive applications.
At each application depth, we fit a $k$-NN classifier and measure its
classification accuracy, which continues to improve beyond the training
depth. The t-SNE projections provide a qualitative view of the same behavior, with
progressively cleaner class clusters. Additional random recurrence therefore
seems to refine the hidden representation even after the classifier already
performs well.

\subsection{Logic Puzzles as Language Modeling}
\label{sec:logic-puzzles-lm}

\begin{table*}[t]
  \centering
  \begin{minipage}[t]{0.485\textwidth}
    \centering
    \caption{Sudoku and Zebra complete-puzzle accuracy.}
    \label{tab:lm-puzzles}
    \small
    \setlength{\tabcolsep}{1.5pt}
    \begin{tabular*}{\linewidth}{@{\extracolsep{\fill}}llccrr@{}}
      \toprule
      Task & Model & $L$ & $T$ & Params. & Acc. (\%) \\
      \midrule
      Sudoku & Dense & 8 & 8 & 42.66M & $85.779 \pm 0.000$ \\
             & RRM & 2 & 16 & 10.78M & $84.523 \pm 0.028$ \\
             & RRM & 4 & 16 & 21.40M & $\mathbf{85.954 \pm 0.131}$ \\
      \addlinespace
      Zebra & Dense & 8 & 8 & 42.89M & $94.980 \pm 0.000$ \\
            & RRM & 2 & 16 & 11.01M & $\mathbf{95.547 \pm 0.103}$ \\
      \bottomrule
    \end{tabular*}
  \end{minipage}\hfill
  \begin{minipage}[t]{0.485\textwidth}
    \centering
    \caption{Maze-Hard results. $\dagger$ evaluates the same RRM with
    $T_{\mathrm{test}}=24$. Baseline results are from
    \citet{jolicoeurmartineau2025morerecursivereasoningtiny}.}
    \label{tab:maze-results}
    \small
    \setlength{\tabcolsep}{3pt}
    \begin{tabular*}{\linewidth}{@{\extracolsep{\fill}}lccrr@{}}
      \toprule
      Model & $L$ & $T$ & Params. & Acc. (\%) \\
      \midrule
      HRM & 8 & 24 & 27M & $74.50$ \\
      TRM-Att & 2 & 30 & 7M & $85.30$ \\
      RRM & 2 & 16 & 7M & $82.90 \pm 0.17$ \\
      RRM$^{\dagger}$ & 2 & 24 & 7M & $\mathbf{87.07 \pm 0.71}$ \\
      \addlinespace[3pt]
      \bottomrule
    \end{tabular*}
  \end{minipage}
\end{table*}

Table~\ref{tab:lm-puzzles} shows that RRM matches or improves upon the
fixed-order dense baselines on Sudoku and Zebra. On Sudoku, the $L=2$ RRM uses
roughly one quarter as many parameters while remaining close to the dense baseline, whereas
$L=4$ exceeds it using about half as many parameters. On Zebra, the $L=2$ RRM
similarly uses roughly one quarter as many parameters and improves accuracy.
The dense baseline results are our reproductions of the models
from \citet{shah2024causallanguagemodelingelicit}.

\subsection{Extreme Reasoning Tasks}

\paragraph{Maze-Hard.}
Table~\ref{tab:maze-results} shows that basic RRM approaches TRM-Att at its
training depth of $T_{\mathrm{train}}=16$. Increasing the same RRM to
$T_{\mathrm{test}}=24$ moves it above TRM-Att, which uses $T=30$ applications,
without retraining or adding parameters. On Maze-Hard, RRM does not use deep
supervision or a training-time halting objective. Unlike the prescribed
recurrent structures of HRM and TRM-Att, it passes a single hidden state
through $L=2$ randomly reused layers.

\begin{wraptable}{r}{0.58\textwidth}
  \vspace{-1.9em}
  \caption{Sudoku-Extreme exact-grid accuracy. HRM, TRM, and TRM+RRM
  use $N_{\mathrm{sup}}=16$, while RRM uses $N_{\mathrm{sup}}=32$.
  $\dagger$ evaluates the same RRM with $T_{\mathrm{test}}=32$ without retraining.
  The TRM row is our reproduction; the published result is $87.40\%$
  \citep{jolicoeurmartineau2025morerecursivereasoningtiny}.}
  \label{tab:sudoku-extreme-results}
  \centering
  \small
  \begin{tabular*}{\linewidth}{@{\extracolsep{\fill}}lcrrr@{}}
    \toprule
    Model & $L$ & $T$ & Params. & Accuracy (\%) \\
    \midrule
    HRM & 8 & 24 & 27M & $55.00$ \\
    TRM & 2 & 42 & 5M & $87.63$ \\
    RRM & 2 & 21 & 5M & $87.22 \pm 0.03$ \\
    TRM+RRM & 2 & 42 & 5M & $89.17 \pm 0.02$ \\
    RRM$^{\dagger}$ & 2 & 32 & 5M & $\mathbf{89.21 \pm 0.03}$ \\
    \bottomrule
  \end{tabular*}
  \vspace{-0.6em}
\end{wraptable}

\paragraph{Sudoku-Extreme.}
Table~\ref{tab:sudoku-extreme-results} shows that RRM closely matches TRM at
the same 5M parameter budget and total depth of 672 layer applications. Rather
than TRM's prescribed recurrent structure, RRM carries a single state through
randomly reused layers. Specifically, RRM uses $T_{\mathrm{train}}=21$
applications over $N_{\mathrm{sup}}=32$ supervision steps, while TRM uses
$T=42$ over $N_{\mathrm{sup}}=16$ steps. TRM+RRM, which changes only TRM's
fixed layer order to random reuse, improves upon both the published TRM result
and our dense reproduction. Increasing basic RRM to
$T_{\mathrm{test}}=32$ produces the strongest result without retraining.

\begin{wraptable}{r}{0.58\textwidth}
  \vspace{-1.9em}
  \caption{ARC-AGI-1 Pass@2 accuracy. Baseline results are from
  \citet{jolicoeurmartineau2025morerecursivereasoningtiny}.
  $\dagger$ evaluates RRM with $N_{\mathrm{sup}}=2$ instead of the 32 used
  during training.}
  \label{tab:arc-agi-1-results}
  \centering
  \small
  \setlength{\tabcolsep}{1.5pt}
  \begin{tabular*}{\linewidth}{@{\extracolsep{\fill}}lccrrr@{}}
    \toprule
    Model & $L$ & $T$ & $T\!\times\!N_{\mathrm{sup}}$ & Params. & Pass@2 (\%) \\
    \midrule
    HRM & 8 & 24 & 384 & 27M & $40.30$ \\
    TRM-Att & 2 & 30 & 480 & 7M & $44.60$ \\
    RRM$^\dagger$ & 2 & 15 & 30 & 7M & $45.79 \pm 0.29$ \\
    RRM & 2 & 15 & 480 & 7M & $\mathbf{47.38 \pm 0.25}$ \\
    \bottomrule
  \end{tabular*}
  \vspace{-0.9em}
\end{wraptable}

\paragraph{ARC-AGI-1.}
Table~\ref{tab:arc-agi-1-results} shows that RRM outperforms the reported TRM
baseline and remains competitive when evaluated with $16\times$ fewer supervision steps at inference time.
Additional details and results are provided in
Appendix~\ref{sec:additional-arc-agi-1}.

On both Sudoku-Extreme and ARC-AGI-1, deep supervision was important during
training.\footnote{An
\href{https://arcprize.org/blog/hrm-analysis}{ARC Prize analysis} likewise
identifies HRM's outer loop as an essential performance driver.} For instance,
RRM memorized the Sudoku-Extreme training data and generalized poorly without
it.

\subsection{Parameter and Compute Efficiency}

\begin{figure*}[t]
  \centering
  \includegraphics[width=\textwidth]{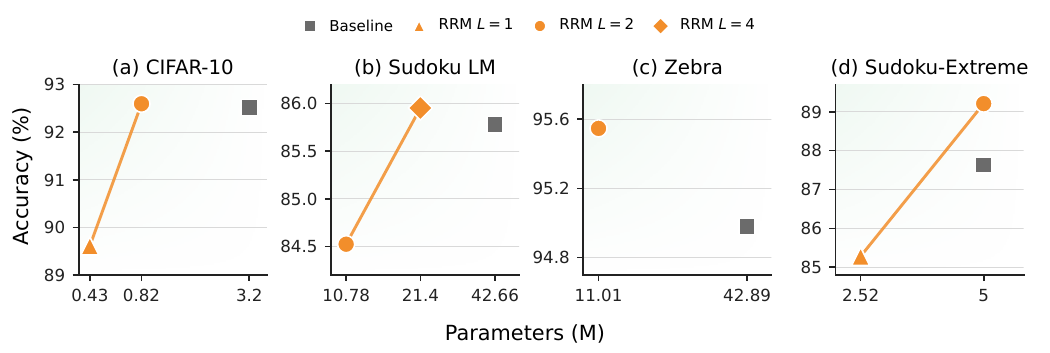}
  \vspace{0.25em}
  \includegraphics[width=\textwidth]{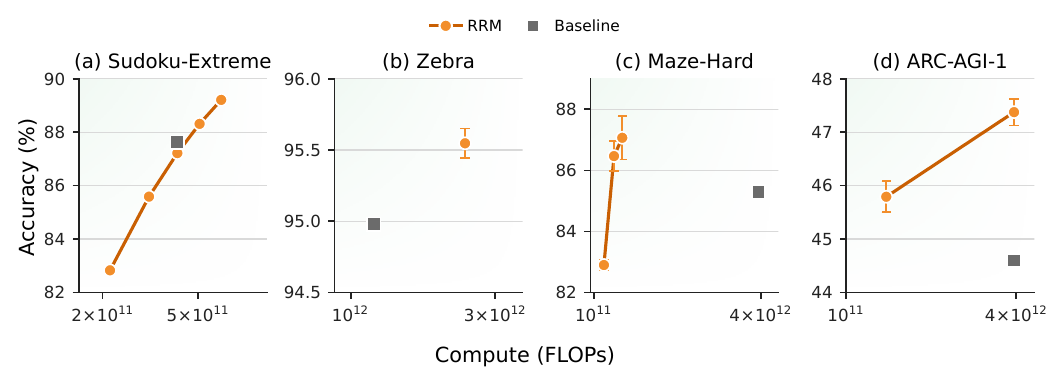}
  \caption{\small Top: Parameter efficiency across four tasks. RRM matches or
  outperforms the corresponding baselines with substantially fewer parameters
  in (a--c) and at a comparable parameter count in (d). Sudoku LM denotes the
  autoregressive Sudoku task in Section~\ref{sec:logic-puzzles-lm}. Bottom:
  Inference compute efficiency across four reasoning tasks. FLOPs are measured
  per evaluation example and include all applications across supervision steps
  when deep supervision is used.
  Overall, RRM consistently improves parameter efficiency and achieves
  competitive or better compute efficiency on three of the four reasoning
  tasks, with the largest gains on Maze-Hard and ARC-AGI-1.}
  \label{fig:efficiency-frontiers}
  \vspace{-1.0em}
\end{figure*}

\paragraph{Parameter efficiency.}
Figure~\ref{fig:efficiency-frontiers} (top) visualizes the parameter efficiency
across four tasks and shows that RRM matches or outperforms the baselines with
substantially fewer parameters. Specifically, $L=2$ uses roughly one quarter
as many parameters on CIFAR-10 and Zebra, while $L=4$ improves Sudoku accuracy
with about half as many.

\newpage
\paragraph{Inference compute efficiency.}
Figure~\ref{fig:efficiency-frontiers} (bottom) shows that RRM exceeds the baselines
on Maze-Hard and ARC-AGI-1 while using only a small fraction of their inference
FLOPs. On Sudoku-Extreme, RRM remains close to TRM at
comparable compute and surpasses it with additional applications.
Interestingly, Zebra instead gains accuracy at higher compute, showing that
parameter efficiency does not necessarily imply compute efficiency.

\begin{wrapfigure}[11]{r}{0.35\textwidth}
  \vspace{-1.5em}
  \centering
  \includegraphics[width=\linewidth,trim={12pt 17pt 5pt 19pt},clip]{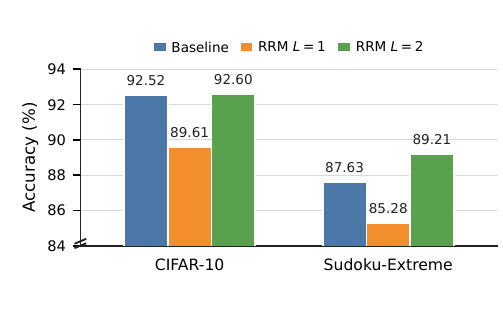}
  \caption{\small Effect of the number of reusable layers. RRM remains
  effective with $L=1$, while $L=2$ matches or exceeds the baselines on both
  tasks.}
  \label{fig:layer-count-comparison}
  \vspace{-0.6em}
\end{wrapfigure}

\paragraph{RRM with \textit{only} one reusable layer.}
To test whether RRM can be reduced to repeatedly applying a single
transformation, we study $L=1$, where one learned function is reused over $T$
applications. This retains recursive computation but removes random layer
selection. The one-layer models perform strongly on both CIFAR-10 and
Sudoku-Extreme, showing that repeatedly applying one function can already learn
useful computation. Under the same setup, however, $L=2$ improves both tasks.
RRM therefore benefits from having multiple reusable functions. Since $L=2$
also adds parameters, this comparison does not isolate random ordering from the
additional model capacity (Figure~\ref{fig:layer-count-comparison}).

\begin{figure*}[t]
  \centering
  \includegraphics[width=\textwidth]{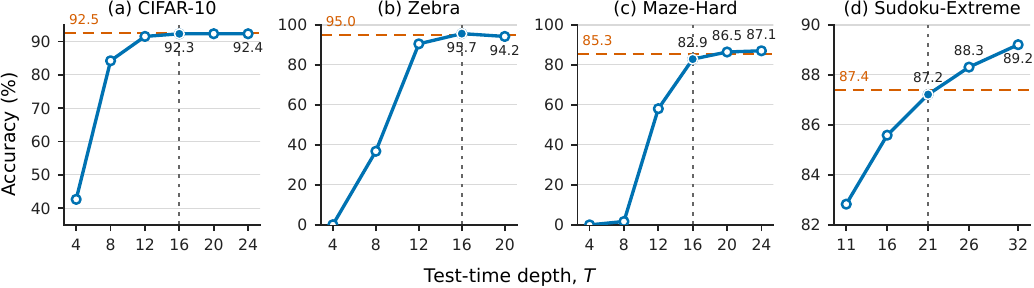}
  \vspace{0.25em}
  \includegraphics[width=\textwidth]{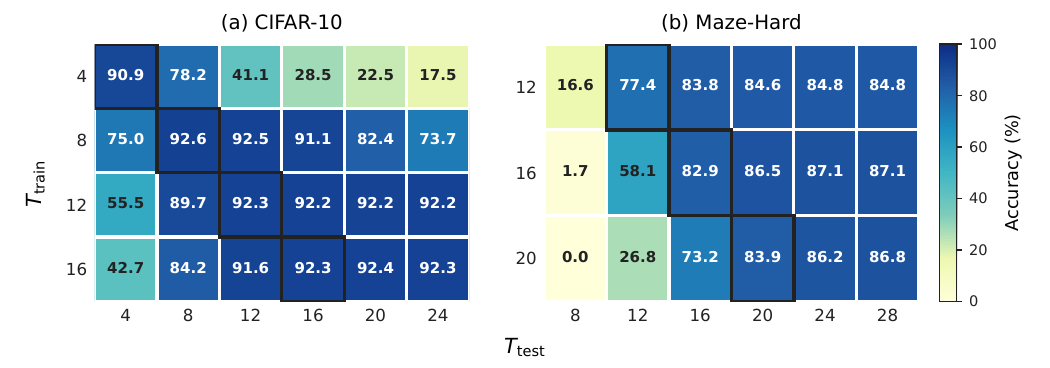}
  \caption{\small Top: Test-time scaling of RRM without retraining or
  adding parameters. Vertical dotted lines mark $T_{\mathrm{train}}$. Horizontal
  dashed lines show the corresponding baselines. Additional test-time
  applications leave performance in (a,b) largely unchanged but improve it in
  (c,d). Bottom: Train-test depth generalization. Rows indicate
  $T_{\mathrm{train}}$ and columns indicate $T_{\mathrm{test}}$; outlined cells
  mark $T_{\mathrm{train}}=T_{\mathrm{test}}$. Across both tasks, performance
  decreases as $T_{\mathrm{test}}$ moves farther below $T_{\mathrm{train}}$.
  When $T_{\mathrm{test}}>T_{\mathrm{train}}$, Maze-Hard improves or plateaus,
  whereas CIFAR-10 plateaus or degrades depending on $T_{\mathrm{train}}$.}
  \label{fig:depth-scaling-combined}
  \vspace{-0.5em}
\end{figure*}

\subsection{Test-Time Scaling}
\label{sec:test-time-results}

Figure~\ref{fig:depth-scaling-combined} (top) compares how the tasks respond when the same
RRM parameters are run for longer. The CIFAR-10 RRM trained with
$T_{\mathrm{train}}=16$ remains stable beyond its training depth, showing robustness to
additional computation rather than a meaningful accuracy gain. Additional
sweeps in Appendix~\ref{sec:additional-cifar10} show that the RRM trained with
$T_{\mathrm{train}}=8$ degrades when extended substantially beyond
its training depth, even with Monte Carlo inference. Zebra follows a similar pattern. 
Its accuracy rises sharply toward the training depth and remains high at the next tested depth, 
but does not improve monotonically as computation is extended further.
We observe a similar pattern on ARC-AGI-1
(Table~\ref{tab:arc-agi-1-ablations}). For the
$T_{\mathrm{train}}=8$, $T_{\mathrm{detach}}=0$ RRM, increasing
$T_{\mathrm{test}}$ from 8 to 15 yields only a marginal gain. For the
$T_{\mathrm{train}}=15$, $T_{\mathrm{detach}}=5$ RRM, increasing
$T_{\mathrm{test}}$ from 15 to 21 provides no gain.

\paragraph{Train-test depth generalization.}
We further visualize test-time depth generalization on CIFAR-10 and Maze-Hard,
where $T_{\mathrm{train}}$ and $T_{\mathrm{test}}$ interact differently
(Figure~\ref{fig:depth-scaling-combined}, bottom). On CIFAR-10, models trained
at shallow depths degrade when extended far beyond $T_{\mathrm{train}}$,
whereas deeper models remain stable across a wider range of
$T_{\mathrm{test}}$. On Maze-Hard, every model improves beyond its training
depth before eventually plateauing. One pattern is common to both tasks.
When $T_{\mathrm{test}}<T_{\mathrm{train}}$, performance decreases as the gap
between them grows, suggesting that models trained for deeper computation may
not tolerate fewer applications at inference. However, understanding this
behavior requires further investigation.

Overall, Maze-Hard and Sudoku-Extreme benefit from additional test-time
computation, with Sudoku-Extreme improving monotonically over all tested
depths. Additional random recurrence at test time may therefore be most
useful for tasks that benefit from longer iterative computation, while gains
plateau or degrade when fewer applications suffice.

\paragraph{Probabilistic test-time scaling.}
\begin{wraptable}[6]{r}{0.47\textwidth}
  \vspace{-2.1em}
  \caption{\small Sudoku-Extreme accuracy with standard inference and PTRM
  test-time scaling.}
  \label{tab:ptrm-results}
  \centering
  \small
  \begin{tabular*}{\linewidth}{@{\extracolsep{\fill}}lcc@{}}
    \toprule
    Model & Standard (\%) & PTRM (\%) \\
    \midrule
    TRM & $87.20$ & $98.43 \pm 0.12$ \\
    TRM+RRM & $88.57 \pm 0.25$ & $98.83 \pm 0.06$ \\
    RRM & $87.20 \pm 0.26$ & $\mathbf{99.50 \pm 0.10}$ \\
    \bottomrule
  \end{tabular*}
  \vspace{-0.01em}
\end{wraptable}

We further evaluate RRM using PTRM's probabilistic test-time scaling
\citep{sghaier2026probabilistictinyrecursivemodel}. PTRM injects Gaussian noise
into the hidden state to produce multiple trajectories, then uses the existing
$Q$ head to choose the solution most likely to be correct
(Appendix~\ref{sec:ptrm-inference-appendix}).

We use the paper's main Sudoku-Extreme
configuration of 100 trajectories, $N_{\mathrm{sup}}=64$ supervision steps,
and $\sigma=0.3$. Because evaluating 100 trajectories per puzzle is expensive,
we use a fixed 1,000-puzzle subset. As shown in
Table~\ref{tab:ptrm-results}, this approach
transfers to all three architectures without retraining and substantially
improves accuracy, with RRM achieving the highest result.

\section{Related Work}
\label{sec:related-work}

RRM is a simple form of parameter-sharing recursion, rather than a replacement for existing recursive or capacity-scaling methods. 
By removing a fixed layer order, it provides a direct way to study how
flexible layer reuse interacts with model capacity, computational depth, and
reasoning. Our experiments provide initial evidence at the tasks and scales
considered, while broader architectures, scales, and reuse strategies remain
open for further study. Below, we first distinguish methods that increase
capacity at roughly fixed computation from methods that increase computation
through parameter sharing, where RRM fits most directly. We then review work on
recursive reasoning, test-time scaling, layer order, and randomized networks.

\subsection{Scaling Capacity without Proportional Increase in Computation}

Capacity-scaling methods increase the number of available parameters while
activating only a subset for each input. Mixtures of local experts introduced
learned specialization through a gating network
\citep{jacobs1991adaptive}, while later conditional-computation methods
generalized this idea to avoid evaluating the entire model
\citep{bengio2013estimatingpropagatinggradientsstochastic,
bengio2016conditionalcomputationneuralnetworks}. Sparsely gated
mixture-of-experts models made this approach practical at scale
\citep{shazeer2017outrageouslylargeneuralnetworks}, and subsequent systems
expanded model capacity while limiting the number of experts applied to each
token \citep{lepikhin2020gshardscalinggiantmodels,
fedus2022switchtransformersscalingtrillion}. Later work improved how tokens and
experts are assigned \citep{zhou2022mixtureofexpertsexpertchoicerouting,
puigcerver2024sparsesoftmixturesexperts}. These methods provide access to a
large parameter pool without a proportional increase in computation. 
In contrast, RRM changes the reuse order of a fixed set of learned layers at a
given computational depth and remains compatible with MoEs, since each
reusable layer could itself be an MoE.

\subsection{Scaling Computation with Shared Parameters}

Parameter-sharing methods instead increase computational depth by applying the
same parameters multiple times. Neural GPUs use a recurrent convolutional
transition to perform multiple
computational steps with shared parameters \citep{kaiser2016neuralgpuslearnalgorithms}.
Universal Transformers apply the same principle to self-attention by repeatedly
applying a shared transition across depth \citep{dehghani2019universaltransformers},
while ALBERT uses cross-layer parameter sharing to prevent the parameter count
from growing with network depth \citep{lan2020albertlitebertselfsupervised}.
More recently, looped and recursive Transformers have been studied for
iterative algorithms, length generalization, parameter-efficient language
modeling, latent reasoning, and adaptive computational depth
\citep{yang2024loopedtransformersbetterlearning,
fan2025loopedtransformerslengthgeneralization,
bae2025relaxedrecursivetransformerseffective,
geiping2025scalingtesttimecomputelatent,
chen2025innerthinkingtransformerleveraging,
bae2025mixtureofrecursionslearningdynamicrecursive,
zhu2026scalinglatentreasoninglooped,
sapunov2026universaltransformersneedmemory,
jeddi2026loopformerelasticdepthloopedtransformers}. Although these approaches
differ in how depth is trained or allocated, they repeatedly apply one
transformation or a fixed sequence of transformations. RRM retains parameter
sharing across depth but samples each application from a pool of learned
layers.

\subsection{Recursive Reasoning and Test-Time Scaling}

Recursive reasoning methods refine an internal state over multiple steps using
shared parameters. Recurrent Relational Networks showed that repeated
relational updates can solve tasks including Sudoku
\citep{palm2018recurrentrelationalnetworks}. Later work found that recurrent
networks trained on easier problems can extrapolate to harder instances by
using additional recurrences at inference, but also identified
\emph{overthinking}, where performance deteriorates when recurrence extends too
far beyond training \citep{schwarzschild2021learnalgorithmgeneralizingeasy,
bansal2022endtoendalgorithmsynthesisrecurrent}. More recently, the Hierarchical
Reasoning Model and Tiny Recursive Model have shown that compact recurrent
networks can solve difficult Sudoku, maze, and abstraction tasks
\citep{wang2025hierarchicalreasoningmodel,
jolicoeurmartineau2025morerecursivereasoningtiny}. RRM instead studies a simpler
form of recursive reasoning, where layers are trained under random reuse and
may learn computations that other models encode through a prescribed recurrent
structure.

Varying the number of recurrent steps also connects recursive reasoning to
test-time scaling. Test-time scaling can allocate additional inference
computation through search and output refinement
\citep{snell2024scalingllmtesttimecompute}. Recursive models can instead scale
latent computation by increasing recurrent depth
\citep{geiping2025scalingtesttimecomputelatent,
zhu2026scalinglatentreasoninglooped}. RRM provides this form of scaling through
randomly reused layers. In the basic RRM, $T$ is set directly and can be
increased during training or at test time without adding learned parameters.
RRM also supports probabilistic test-time scaling without retraining
\citep{sghaier2026probabilistictinyrecursivemodel}.

\subsection{Layer Order and Randomized Networks}

Many methods alter the computational path through a network.
Early-exit and layer-skipping methods shorten a fixed sequence of layers while
preserving its relative order
\citep{teerapittayanon2017branchynetfastinferenceearly,
wang2018skipnetlearningdynamicrouting,
wu2019blockdropdynamicinferencepaths}. PathNet and Routing Networks select
task-specific paths through reusable modules
\citep{fernando2017pathnetevolutionchannelsgradient,
rosenbaum2017routingnetworksadaptiveselection}, while Modular Networks and
Distributed Neural Architectures learn input- or token-dependent compositions
that may change module order
\citep{kirsch2018modularnetworkslearningdecompose,
cowsik2025distributedneuralarchitectures}. Other work searches at test time for
ways to skip or repeat contiguous blocks of a pretrained language model
\citep{li2025skiplayerloopit}. These approaches either retain a constrained
order, specialize routes at the task level, or learn discrete routing decisions,
whose non-differentiability complicates end-to-end training.

Randomized-network methods instead introduce stochasticity without learning an
explicit routing policy. LayerShuffle randomly permutes all layers once per
batch to make a full-size Vision Transformer robust to reordering
\citep{LayerShuffle}, whereas RRM samples with replacement per example and
step, enabling $T\ne L$, parameter-efficient recursion, and inference-depth
scaling. Dropout randomly masks individual activations
\citep{srivastava14a}. At the level of depth, stochastic depth and LayerDrop
randomly omit entire layers, while Swapout and FractalNet sample paths through
larger fixed architectures
\citep{huang2016deepnetworksstochasticdepth,
fan2019reducingtransformerdepthdemand,
singh2016swapoutlearningensembledeep,
larsson2017fractalnetultradeepneuralnetworks}. These methods vary which fixed
components are active while preserving their relative order. At the level of
network topology, randomly wired neural networks sample graph connectivity
before training, after which the sampled graph defines the architecture
\citep{xie2019exploringrandomlywiredneural}. RRM instead samples a layer at each
step, allowing arbitrary order and reuse without learning or searching for a
route.

\section{Conclusion and Future Work}
We introduced the Random Recursive Model (RRM), which replaces fixed-order
recurrence with random reuse of a small set of learned layers. RRM matches or
exceeds the corresponding baselines, often with
substantial reductions in parameter count. Separating the number of learned
layers from the
number of applications also allows RRM to scale computation without increasing
model size. Additional inference time applications improve performance on Maze-Hard and
Sudoku-Extreme, while gains plateau or eventually decline on tasks that require
less computation. Monte Carlo inference and probabilistic test-time scaling
provide further improvements without retraining.
Together, these results suggest that random layer reuse acts as an effective
form of regularization, improving generalization while
mitigating overparameterization.

Future work should study how the sampling distribution and the diversity of
sampled sequences affect learning. We also studied learned sampling policies, but did not obtain useful gains (Appendix~\ref{sec:learned-ordering-appendix}). 
Making learned ordering work remains an important direction, since an adaptive policy
could reduce computation through earlier exiting and improve parameter
efficiency by using a smaller subset of reusable layers. Future work should
also study how scaling the number of layers and applications behaves in larger
models and tasks. More fundamentally, it remains
unclear why random layer reuse is effective, and studying its theoretical
foundations is left for future work. More
broadly, these findings may guide the design of networks that retain flexible
layer reuse while being even more computationally efficient.

\newpage
%



%


\subsubsection*{Acknowledgments}
This work would not have been possible without the feedback and suggestions of many
colleagues and friends at Mila. We thank Alex Hern\'andez-Garc\'ia, Lucas Maes,
Mohsin Hasan, Hafez Ghaemi, Gauthier Gidel, F{\i}rat \"Oncel, and Jay Gala for
their helpful comments and suggestions. We are especially grateful to Olexa
Bilaniuk and the other members of the IDT team for their technical support with
the Mila and DRAC computing clusters. We also thank Cem Subakan for providing
additional compute resources.
Finally, we thank the authors of the datasets and baseline implementations used
in this work for making them publicly available.
The research was enabled in part by computational resources provided by the
Digital Research Alliance of Canada (\url{https://alliancecan.ca}) and Mila
(\url{https://mila.quebec}).

\subsubsection*{Broader Impacts and AI Safety}
RRM composes reusable modules by passing a shared hidden state through a sampled
order. Although our experiments use layers in small task-specific models, these
modules could be replaced by larger agents, making the hidden state an implicit
channel between them. Random or learned ordering may improve flexible reasoning,
but may also produce difficult-to-monitor coordination or unsafe behavior.
Future work that extends RRM to agents, learns ordering policies, or searches
over network distributions should therefore constrain possible interactions,
record routing decisions, and evaluate safety before deployment.

\bibliography{iclr2027_conference}

@misc{vaswani2023attentionneed,
  title         = {Attention Is All You Need},
  author        = {Ashish Vaswani and Noam Shazeer and Niki Parmar and Jakob Uszkoreit and Llion Jones and Aidan N. Gomez and Lukasz Kaiser and Illia Polosukhin},
  year          = {2023},
  eprint        = {1706.03762},
  archiveprefix = {arXiv},
  primaryclass  = {cs.CL},
  url           = {https://arxiv.org/abs/1706.03762}
}

@misc{dosovitskiy2021image,
  title         = {An Image is Worth 16x16 Words: Transformers for Image Recognition at Scale},
  author        = {Alexey Dosovitskiy and Lucas Beyer and Alexander Kolesnikov and Dirk Weissenborn and Xiaohua Zhai and Thomas Unterthiner and Mostafa Dehghani and Matthias Minderer and Georg Heigold and Sylvain Gelly and Jakob Uszkoreit and Neil Houlsby},
  year          = {2021},
  eprint        = {2010.11929},
  archiveprefix = {arXiv},
  primaryclass  = {cs.CV},
  url           = {https://arxiv.org/abs/2010.11929}
}

@misc{cubuk2019autoaugment,
  title         = {AutoAugment: Learning Augmentation Policies from Data},
  author        = {Ekin D. Cubuk and Barret Zoph and Dandelion Mane and Vijay Vasudevan and Quoc V. Le},
  year          = {2019},
  eprint        = {1805.09501},
  archiveprefix = {arXiv},
  primaryclass  = {cs.CV},
  url           = {https://arxiv.org/abs/1805.09501}
}

@misc{chollet2019measure,
  title         = {On the Measure of Intelligence},
  author        = {François Chollet},
  year          = {2019},
  eprint        = {1911.01547},
  archiveprefix = {arXiv},
  primaryclass  = {cs.AI},
  url           = {https://arxiv.org/abs/1911.01547}
}

@misc{dehghani2019universaltransformers,
  title         = {Universal Transformers},
  author        = {Mostafa Dehghani and Stephan Gouws and Oriol Vinyals and Jakob Uszkoreit and Łukasz Kaiser},
  year          = {2019},
  eprint        = {1807.03819},
  archiveprefix = {arXiv},
  primaryclass  = {cs.CL},
  url           = {https://arxiv.org/abs/1807.03819}
}

@misc{yang2024loopedtransformersbetterlearning,
  title         = {Looped Transformers are Better at Learning Learning Algorithms},
  author        = {Liu Yang and Kangwook Lee and Robert Nowak and Dimitris Papailiopoulos},
  year          = {2024},
  eprint        = {2311.12424},
  archiveprefix = {arXiv},
  primaryclass  = {cs.LG},
  url           = {https://arxiv.org/abs/2311.12424}
}

@misc{geiping2025scalingtesttimecomputelatent,
  title         = {Scaling up Test-Time Compute with Latent Reasoning: A Recurrent Depth Approach},
  author        = {Jonas Geiping and Sean McLeish and Neel Jain and John Kirchenbauer and Siddharth Singh and Brian R. Bartoldson and Bhavya Kailkhura and Abhinav Bhatele and Tom Goldstein},
  year          = {2025},
  eprint        = {2502.05171},
  archiveprefix = {arXiv},
  primaryclass  = {cs.LG},
  url           = {https://arxiv.org/abs/2502.05171}
}

@misc{wang2025hierarchicalreasoningmodel,
  title         = {Hierarchical Reasoning Model},
  author        = {Guan Wang and Jin Li and Yuhao Sun and Xing Chen and Changling Liu and Yue Wu and Meng Lu and Sen Song and Yasin Abbasi Yadkori},
  year          = {2025},
  eprint        = {2506.21734},
  archiveprefix = {arXiv},
  primaryclass  = {cs.AI},
  url           = {https://arxiv.org/abs/2506.21734}
}

@misc{jolicoeurmartineau2025morerecursivereasoningtiny,
  title         = {Less is More: Recursive Reasoning with Tiny Networks},
  author        = {Alexia Jolicoeur-Martineau},
  year          = {2025},
  eprint        = {2510.04871},
  archiveprefix = {arXiv},
  primaryclass  = {cs.LG},
  url           = {https://arxiv.org/abs/2510.04871}
}

@misc{bae2025relaxedrecursivetransformerseffective,
  title         = {Relaxed Recursive Transformers: Effective Parameter Sharing with Layer-wise LoRA},
  author        = {Sangmin Bae and Adam Fisch and Hrayr Harutyunyan and Ziwei Ji and Seungyeon Kim and Tal Schuster},
  year          = {2025},
  eprint        = {2410.20672},
  archiveprefix = {arXiv},
  primaryclass  = {cs.CL},
  url           = {https://arxiv.org/abs/2410.20672}
}

@misc{zhu2026scalinglatentreasoninglooped,
  title         = {Scaling Latent Reasoning via Looped Language Models},
  author        = {Rui-Jie Zhu and Zixuan Wang and Kai Hua and Tianyu Zhang and Ziniu Li and Haoran Que and Boyi Wei and Zixin Wen and Fan Yin and He Xing and Lu Li and Jiajun Shi and Kaijing Ma and Shanda Li and Taylor Kergan and Andrew Smith and Xingwei Qu and Mude Hui and Bohong Wu and Qiyang Min and Hongzhi Huang and Xun Zhou and Wei Ye and Jiaheng Liu and Jian Yang and Yunfeng Shi and Chenghua Lin and Enduo Zhao and Tianle Cai and Ge Zhang and Wenhao Huang and Yoshua Bengio and Jason Eshraghian},
  year          = {2026},
  eprint        = {2510.25741},
  archiveprefix = {arXiv},
  primaryclass  = {cs.CL},
  url           = {https://arxiv.org/abs/2510.25741}
}

@misc{sghaier2026probabilistictinyrecursivemodel,
  title         = {Probabilistic Tiny Recursive Model},
  author        = {Amin Sghaier and Ali Parviz and Alexia Jolicoeur-Martineau},
  year          = {2026},
  eprint        = {2605.19943},
  archiveprefix = {arXiv},
  primaryclass  = {cs.AI},
  url           = {https://arxiv.org/abs/2605.19943}
}

@misc{snell2024scalingllmtesttimecompute,
  title         = {Scaling LLM Test-Time Compute Optimally can be More Effective than Scaling Model Parameters},
  author        = {Charlie Snell and Jaehoon Lee and Kelvin Xu and Aviral Kumar},
  year          = {2024},
  eprint        = {2408.03314},
  archiveprefix = {arXiv},
  primaryclass  = {cs.LG},
  url           = {https://arxiv.org/abs/2408.03314}
}

@techreport{krizhevsky2009learning,
  author                      = {Krizhevsky, Alex and Hinton, Geoffrey},
  address                     = {Toronto, Ontario},
  institution                 = {University of Toronto},
  number                      = {0},
  publisher                   = {Technical report, University of Toronto},
  title                       = {Learning multiple layers of features from tiny images},
  year                        = {2009},
  title_with_no_special_chars = {Learning multiple layers of features from tiny images},
  url                         = {https://www.cs.toronto.edu/~kriz/learning-features-2009-TR.pdf}
}

@misc{shah2024causallanguagemodelingelicit,
  title         = {Causal Language Modeling Can Elicit Search and Reasoning Capabilities on Logic Puzzles},
  author        = {Kulin Shah and Nishanth Dikkala and Xin Wang and Rina Panigrahy},
  year          = {2024},
  eprint        = {2409.10502},
  archiveprefix = {arXiv},
  primaryclass  = {cs.LG},
  url           = {https://arxiv.org/abs/2409.10502}
}

@article{vandermaaten08a,
  author  = {Laurens van der Maaten and Geoffrey Hinton},
  title   = {Visualizing Data using t-SNE},
  journal = {Journal of Machine Learning Research},
  year    = {2008},
  volume  = {9},
  number  = {86},
  pages   = {2579--2605},
  url     = {http://jmlr.org/papers/v9/vandermaaten08a.html}
}

@article{jacobs1991adaptive,
  author   = {Jacobs, Robert A. and Jordan, Michael I. and Nowlan, Steven J. and Hinton, Geoffrey E.},
  journal  = {Neural Computation},
  title    = {Adaptive Mixtures of Local Experts},
  year     = {1991},
  volume   = {3},
  number   = {1},
  pages    = {79-87},
  doi      = {10.1162/neco.1991.3.1.79}
}

@misc{bengio2013estimatingpropagatinggradientsstochastic,
  title         = {Estimating or Propagating Gradients Through Stochastic Neurons for Conditional Computation},
  author        = {Yoshua Bengio and Nicholas Léonard and Aaron Courville},
  year          = {2013},
  eprint        = {1308.3432},
  archiveprefix = {arXiv},
  primaryclass  = {cs.LG},
  url           = {https://arxiv.org/abs/1308.3432}
}

@misc{bengio2016conditionalcomputationneuralnetworks,
  title         = {Conditional Computation in Neural Networks for faster models},
  author        = {Emmanuel Bengio and Pierre-Luc Bacon and Joelle Pineau and Doina Precup},
  year          = {2016},
  eprint        = {1511.06297},
  archiveprefix = {arXiv},
  primaryclass  = {cs.LG},
  url           = {https://arxiv.org/abs/1511.06297}
}

@misc{shazeer2017outrageouslylargeneuralnetworks,
  title         = {Outrageously Large Neural Networks: The Sparsely-Gated Mixture-of-Experts Layer},
  author        = {Noam Shazeer and Azalia Mirhoseini and Krzysztof Maziarz and Andy Davis and Quoc Le and Geoffrey Hinton and Jeff Dean},
  year          = {2017},
  eprint        = {1701.06538},
  archiveprefix = {arXiv},
  primaryclass  = {cs.LG},
  url           = {https://arxiv.org/abs/1701.06538}
}

@misc{lepikhin2020gshardscalinggiantmodels,
  title         = {GShard: Scaling Giant Models with Conditional Computation and Automatic Sharding},
  author        = {Dmitry Lepikhin and HyoukJoong Lee and Yuanzhong Xu and Dehao Chen and Orhan Firat and Yanping Huang and Maxim Krikun and Noam Shazeer and Zhifeng Chen},
  year          = {2020},
  eprint        = {2006.16668},
  archiveprefix = {arXiv},
  primaryclass  = {cs.CL},
  url           = {https://arxiv.org/abs/2006.16668}
}

@misc{fedus2022switchtransformersscalingtrillion,
  title         = {Switch Transformers: Scaling to Trillion Parameter Models with Simple and Efficient Sparsity},
  author        = {William Fedus and Barret Zoph and Noam Shazeer},
  year          = {2022},
  eprint        = {2101.03961},
  archiveprefix = {arXiv},
  primaryclass  = {cs.LG},
  url           = {https://arxiv.org/abs/2101.03961}
}

@misc{zhou2022mixtureofexpertsexpertchoicerouting,
  title         = {Mixture-of-Experts with Expert Choice Routing},
  author        = {Yanqi Zhou and Tao Lei and Hanxiao Liu and Nan Du and Yanping Huang and Vincent Zhao and Andrew Dai and Zhifeng Chen and Quoc Le and James Laudon},
  year          = {2022},
  eprint        = {2202.09368},
  archiveprefix = {arXiv},
  primaryclass  = {cs.LG},
  url           = {https://arxiv.org/abs/2202.09368}
}

@misc{puigcerver2024sparsesoftmixturesexperts,
  title         = {From Sparse to Soft Mixtures of Experts},
  author        = {Joan Puigcerver and Carlos Riquelme and Basil Mustafa and Neil Houlsby},
  year          = {2024},
  eprint        = {2308.00951},
  archiveprefix = {arXiv},
  primaryclass  = {cs.LG},
  url           = {https://arxiv.org/abs/2308.00951}
}

@misc{graves2017adaptivecomputationtimerecurrent,
  title         = {Adaptive Computation Time for Recurrent Neural Networks},
  author        = {Alex Graves},
  year          = {2017},
  eprint        = {1603.08983},
  archiveprefix = {arXiv},
  primaryclass  = {cs.NE},
  url           = {https://arxiv.org/abs/1603.08983}
}

@misc{kaiser2016neuralgpuslearnalgorithms,
  title         = {Neural GPUs Learn Algorithms},
  author        = {Łukasz Kaiser and Ilya Sutskever},
  year          = {2016},
  eprint        = {1511.08228},
  archiveprefix = {arXiv},
  primaryclass  = {cs.LG},
  url           = {https://arxiv.org/abs/1511.08228}
}

@misc{lan2020albertlitebertselfsupervised,
  title         = {ALBERT: A Lite BERT for Self-supervised Learning of Language Representations},
  author        = {Zhenzhong Lan and Mingda Chen and Sebastian Goodman and Kevin Gimpel and Piyush Sharma and Radu Soricut},
  year          = {2020},
  eprint        = {1909.11942},
  archiveprefix = {arXiv},
  primaryclass  = {cs.CL},
  url           = {https://arxiv.org/abs/1909.11942}
}

@misc{banino2021pondernetlearningponder,
  title         = {PonderNet: Learning to Ponder},
  author        = {Andrea Banino and Jan Balaguer and Charles Blundell},
  year          = {2021},
  eprint        = {2107.05407},
  archiveprefix = {arXiv},
  primaryclass  = {cs.LG},
  url           = {https://arxiv.org/abs/2107.05407}
}

@misc{bae2025mixtureofrecursionslearningdynamicrecursive,
  title         = {Mixture-of-Recursions: Learning Dynamic Recursive Depths for Adaptive Token-Level Computation},
  author        = {Sangmin Bae and Yujin Kim and Reza Bayat and Sungnyun Kim and Jiyoun Ha and Tal Schuster and Adam Fisch and Hrayr Harutyunyan and Ziwei Ji and Aaron Courville and Se-Young Yun},
  year          = {2025},
  eprint        = {2507.10524},
  archiveprefix = {arXiv},
  primaryclass  = {cs.CL},
  url           = {https://arxiv.org/abs/2507.10524}
}

@misc{fan2025loopedtransformerslengthgeneralization,
  title         = {Looped Transformers for Length Generalization},
  author        = {Ying Fan and Yilun Du and Kannan Ramchandran and Kangwook Lee},
  year          = {2025},
  eprint        = {2409.15647},
  archiveprefix = {arXiv},
  primaryclass  = {cs.LG},
  url           = {https://arxiv.org/abs/2409.15647}
}

@misc{chen2025innerthinkingtransformerleveraging,
  title         = {Inner Thinking Transformer: Leveraging Dynamic Depth Scaling to Foster Adaptive Internal Thinking},
  author        = {Yilong Chen and Junyuan Shang and Zhenyu Zhang and Yanxi Xie and Jiawei Sheng and Tingwen Liu and Shuohuan Wang and Yu Sun and Hua Wu and Haifeng Wang},
  year          = {2025},
  eprint        = {2502.13842},
  archiveprefix = {arXiv},
  primaryclass  = {cs.CL},
  url           = {https://arxiv.org/abs/2502.13842}
}

@misc{li2025skiplayerloopit,
  title         = {Skip a Layer or Loop it? Test-Time Depth Adaptation of Pretrained LLMs},
  author        = {Ziyue Li and Yang Li and Tianyi Zhou},
  year          = {2025},
  eprint        = {2507.07996},
  archiveprefix = {arXiv},
  primaryclass  = {cs.LG},
  url           = {https://arxiv.org/abs/2507.07996}
}

@misc{teerapittayanon2017branchynetfastinferenceearly,
  title         = {BranchyNet: Fast Inference via Early Exiting from Deep Neural Networks},
  author        = {Surat Teerapittayanon and Bradley McDanel and H. T. Kung},
  year          = {2017},
  eprint        = {1709.01686},
  archiveprefix = {arXiv},
  primaryclass  = {cs.NE},
  url           = {https://arxiv.org/abs/1709.01686}
}

@misc{cowsik2025distributedneuralarchitectures,
  title         = {Towards Distributed Neural Architectures},
  author        = {Aditya Cowsik and Tianyu He and Andrey Gromov},
  year          = {2025},
  eprint        = {2506.22389},
  archiveprefix = {arXiv},
  primaryclass  = {cs.LG},
  url           = {https://arxiv.org/abs/2506.22389}
}

@misc{sapunov2026universaltransformersneedmemory,
  title         = {Universal Transformers Need Memory: Depth-State Trade-offs in Adaptive Recursive Reasoning},
  author        = {Grigory Sapunov},
  year          = {2026},
  eprint        = {2604.21999},
  archiveprefix = {arXiv},
  primaryclass  = {cs.LG},
  url           = {https://arxiv.org/abs/2604.21999}
}

@misc{jeddi2026loopformerelasticdepthloopedtransformers,
  title         = {LoopFormer: Elastic-Depth Looped Transformers for Latent Reasoning via Shortcut Modulation},
  author        = {Ahmadreza Jeddi and Marco Ciccone and Babak Taati},
  year          = {2026},
  eprint        = {2602.11451},
  archiveprefix = {arXiv},
  primaryclass  = {cs.CL},
  url           = {https://arxiv.org/abs/2602.11451}
}

@misc{fernando2017pathnetevolutionchannelsgradient,
  title         = {PathNet: Evolution Channels Gradient Descent in Super Neural Networks},
  author        = {Chrisantha Fernando and Dylan Banarse and Charles Blundell and Yori Zwols and David Ha and Andrei A. Rusu and Alexander Pritzel and Daan Wierstra},
  year          = {2017},
  eprint        = {1701.08734},
  archiveprefix = {arXiv},
  primaryclass  = {cs.NE},
  url           = {https://arxiv.org/abs/1701.08734}
}

@misc{rosenbaum2017routingnetworksadaptiveselection,
  title         = {Routing Networks: Adaptive Selection of Non-linear Functions for Multi-Task Learning},
  author        = {Clemens Rosenbaum and Tim Klinger and Matthew Riemer},
  year          = {2017},
  eprint        = {1711.01239},
  archiveprefix = {arXiv},
  primaryclass  = {cs.LG},
  url           = {https://arxiv.org/abs/1711.01239}
}

@misc{kirsch2018modularnetworkslearningdecompose,
  title         = {Modular Networks: Learning to Decompose Neural Computation},
  author        = {Louis Kirsch and Julius Kunze and David Barber},
  year          = {2018},
  eprint        = {1811.05249},
  archiveprefix = {arXiv},
  primaryclass  = {cs.LG},
  url           = {https://arxiv.org/abs/1811.05249}
}

@misc{wang2018skipnetlearningdynamicrouting,
  title         = {SkipNet: Learning Dynamic Routing in Convolutional Networks},
  author        = {Xin Wang and Fisher Yu and Zi-Yi Dou and Trevor Darrell and Joseph E. Gonzalez},
  year          = {2018},
  eprint        = {1711.09485},
  archiveprefix = {arXiv},
  primaryclass  = {cs.CV},
  url           = {https://arxiv.org/abs/1711.09485}
}

@misc{wu2019blockdropdynamicinferencepaths,
  title         = {BlockDrop: Dynamic Inference Paths in Residual Networks},
  author        = {Zuxuan Wu and Tushar Nagarajan and Abhishek Kumar and Steven Rennie and Larry S. Davis and Kristen Grauman and Rogerio Feris},
  year          = {2019},
  eprint        = {1711.08393},
  archiveprefix = {arXiv},
  primaryclass  = {cs.CV},
  url           = {https://arxiv.org/abs/1711.08393}
}

@misc{huang2016deepnetworksstochasticdepth,
  title         = {Deep Networks with Stochastic Depth},
  author        = {Gao Huang and Yu Sun and Zhuang Liu and Daniel Sedra and Kilian Weinberger},
  year          = {2016},
  eprint        = {1603.09382},
  archiveprefix = {arXiv},
  primaryclass  = {cs.LG},
  url           = {https://arxiv.org/abs/1603.09382}
}

@misc{singh2016swapoutlearningensembledeep,
  title         = {Swapout: Learning an ensemble of deep architectures},
  author        = {Saurabh Singh and Derek Hoiem and David Forsyth},
  year          = {2016},
  eprint        = {1605.06465},
  archiveprefix = {arXiv},
  primaryclass  = {cs.CV},
  url           = {https://arxiv.org/abs/1605.06465}
}

@misc{larsson2017fractalnetultradeepneuralnetworks,
  title         = {FractalNet: Ultra-Deep Neural Networks without Residuals},
  author        = {Gustav Larsson and Michael Maire and Gregory Shakhnarovich},
  year          = {2017},
  eprint        = {1605.07648},
  archiveprefix = {arXiv},
  primaryclass  = {cs.CV},
  url           = {https://arxiv.org/abs/1605.07648}
}

@misc{fan2019reducingtransformerdepthdemand,
  title         = {Reducing Transformer Depth on Demand with Structured Dropout},
  author        = {Angela Fan and Edouard Grave and Armand Joulin},
  year          = {2019},
  eprint        = {1909.11556},
  archiveprefix = {arXiv},
  primaryclass  = {cs.LG},
  url           = {https://arxiv.org/abs/1909.11556}
}

@misc{xie2019exploringrandomlywiredneural,
  title         = {Exploring Randomly Wired Neural Networks for Image Recognition},
  author        = {Saining Xie and Alexander Kirillov and Ross Girshick and Kaiming He},
  year          = {2019},
  eprint        = {1904.01569},
  archiveprefix = {arXiv},
  primaryclass  = {cs.CV},
  url           = {https://arxiv.org/abs/1904.01569}
}

@misc{palm2018recurrentrelationalnetworks,
  title         = {Recurrent Relational Networks},
  author        = {Rasmus Berg Palm and Ulrich Paquet and Ole Winther},
  year          = {2018},
  eprint        = {1711.08028},
  archiveprefix = {arXiv},
  primaryclass  = {cs.AI},
  url           = {https://arxiv.org/abs/1711.08028}
}

@misc{schwarzschild2021learnalgorithmgeneralizingeasy,
  title         = {Can You Learn an Algorithm? Generalizing from Easy to Hard Problems with Recurrent Networks},
  author        = {Avi Schwarzschild and Eitan Borgnia and Arjun Gupta and Furong Huang and Uzi Vishkin and Micah Goldblum and Tom Goldstein},
  year          = {2021},
  eprint        = {2106.04537},
  archiveprefix = {arXiv},
  primaryclass  = {cs.LG},
  url           = {https://arxiv.org/abs/2106.04537}
}

@misc{bansal2022endtoendalgorithmsynthesisrecurrent,
  title         = {End-to-end Algorithm Synthesis with Recurrent Networks: Logical Extrapolation Without Overthinking},
  author        = {Arpit Bansal and Avi Schwarzschild and Eitan Borgnia and Zeyad Emam and Furong Huang and Micah Goldblum and Tom Goldstein},
  year          = {2022},
  eprint        = {2202.05826},
  archiveprefix = {arXiv},
  primaryclass  = {cs.LG},
  url           = {https://arxiv.org/abs/2202.05826}
}

@article{srivastava14a,
  author  = {Nitish Srivastava and Geoffrey Hinton and Alex Krizhevsky and Ilya Sutskever and Ruslan Salakhutdinov},
  title   = {Dropout: A Simple Way to Prevent Neural Networks from Overfitting},
  journal = {Journal of Machine Learning Research},
  year    = {2014},
  volume  = {15},
  number  = {56},
  pages   = {1929--1958},
  url     = {http://jmlr.org/papers/v15/srivastava14a.html}
}

@article{williams1992simple,
  title     = {Simple statistical gradient-following algorithms for connectionist reinforcement learning},
  author    = {Williams, Ronald J},
  journal   = {Machine learning},
  volume    = {8},
  number    = {3},
  pages     = {229--256},
  year      = {1992},
  publisher = {Springer}
}

@misc{LayerShuffle,
  title         = {LayerShuffle: Enhancing Robustness in Vision Transformers by Randomizing Layer Execution Order},
  author        = {Matthias Freiberger and Peter Kun and Anders Sundnes Løvlie and Sebastian Risi},
  year          = {2024},
  eprint        = {2407.04513},
  archiveprefix = {arXiv},
  primaryclass  = {cs.CV},
  url           = {https://arxiv.org/abs/2407.04513}
}
\bibliographystyle{iclr2027_conference}
\newpage
\appendix
\section{Additional Details}

\subsection{Deep Supervision}
\label{sec:deep-supervision-appendix}

We observe that deep supervision \citep{wang2025hierarchicalreasoningmodel} is helpful
for Sudoku-Extreme and ARC-AGI-1. For both tasks, it helps with training
optimization and seems to be a main driver of performance.\footnote{The ARC Prize
analysis also found that HRM's outer refinement loop drove substantial gains on
ARC-AGI.\url{https://arcprize.org/blog/hrm-analysis}} Like other recursive
networks, RRM appears to benefit from deep supervision while retaining simpler,
more flexible layer reuse.

In deep supervision, a single hidden state is carried across supervision steps. Let
$h_t^{(j)}$ denote the state after application $t$ of supervision step $j$,
for $j=1,\ldots,N_{\mathrm{sup}}$ and $t=0,\ldots,T-1$. Then
\[
  h_0^{(1)}=E(x), \qquad
  h_0^{(j)}=\operatorname{sg}\!\left(h_T^{(j-1)}\right), \qquad
  h_{t+1}^{(j)}=f_{r_t^{(j)}}\!\left(h_t^{(j)},E(x)\right),
\]
where the second expression applies for $j>1$ and $\operatorname{sg}$ stops
the gradient. A task loss is applied after each supervision step. The next
step therefore continues from the preceding state, but gradients do not cross
between supervision steps.

Deep supervision also uses early exiting, a popular method in earlier works on
conditional computation, including
\citet{graves2017adaptivecomputationtimerecurrent,banino2021pondernetlearningponder}.
We observe that this is very helpful for probabilistic test-time scaling
\citep{sghaier2026probabilistictinyrecursivemodel}, since it uses the Q head to
select the best candidate (Appendix~\ref{sec:ptrm-inference-appendix}).

\subsection{Experimental Details}
\label{sec:experimental-details-appendix}

\paragraph{CIFAR-10.}
The custom Vision Transformer divides each $32\times32$ image into
$4\times4$ patches and uses post-normalized Transformer blocks with hidden
width 256, eight attention heads, an MLP ratio of one, and no dropout. The
dense baseline contains eight distinct blocks, while RRM uses the same block
design with one or two reusable layers. All models are trained from scratch
for 800 epochs with batch size 512 using Adam, an initial learning rate of
$10^{-3}$, weight decay $5\times10^{-5}$, label smoothing 0.1, and a
learning-rate schedule with 10\% warmup, a 40\% constant phase, and 50\%
linear decay. Training uses random crops, horizontal flips, and the CIFAR-10 AutoAugment policy
\citep{cubuk2019autoaugment}. We reserve a stratified 20\% of the official
training set for validation, select checkpoints by validation accuracy, and
report top-1 accuracy on the official test set.

\paragraph{Sudoku and Zebra.}
Following \citet{shah2024causallanguagemodelingelicit}, we use the released
solver-ordered datasets and eight-layer causal Transformer configuration, with
hidden width 576, eight attention heads, MLP width 3456, and dropout 0.2. All
models are trained for 320,000 optimizer steps. For Sudoku, we use the complete
released training and test sets. For Zebra, we use the complete released
training set and a fixed subset of 15,000 released test puzzles, matching the
evaluation-set size used by
\citet{shah2024causallanguagemodelingelicit}.

\paragraph{Maze-Hard and Sudoku-Extreme.}
We use the Maze-Hard and Sudoku-Extreme datasets introduced by HRM
\citep{wang2025hierarchicalreasoningmodel} and follow the corresponding setups
used by TRM \citep{jolicoeurmartineau2025morerecursivereasoningtiny}. Maze-Hard
contains fixed sets of 1,000 training and 1,000 test mazes. For
Sudoku-Extreme, we use 1,000 training puzzles with 1,000 augmentations per
puzzle and the provided test split. Both backbones have hidden width 512.
Maze-Hard uses non-causal Transformer blocks with eight attention heads and an
MLP expansion ratio of four, while Sudoku-Extreme uses TRM's position- and
channel-mixing blocks with the same expansion ratio. Neither dataset provides
a separate validation split, so the provided test splits are also used for
periodic checkpoint evaluation.

\paragraph{How we compute $T$ for the baselines.}
For HRM, $T$ counts every layer application in both recurrent modules, giving
$T=2(2\times4+4)=24$, from two high-level cycles, two low-level cycles 
per high-level cycle, and four layers in each module. 
For TRM, each of its three recursion cycles performs the low-level updates 
followed by one answer update through the same two-layer network, 
giving $T=3(4+1)\times2=30$ on Maze-Hard and ARC-AGI-1 and
$T=3(6+1)\times2=42$ on Sudoku-Extreme.

\paragraph{Gradient truncation.}
For RRM, we can truncate gradients within a sequence of random recursive
applications. We observe that this improves optimization on Sudoku-Extreme
while reducing training computation. Given $T$ applications, we truncate
gradients through the first $T_{\mathrm{detach}}$ applications. For the
reported Sudoku-Extreme results, we use $T_{\mathrm{detach}}=7$ out of
$T_{\mathrm{train}}=21$ applications.

\paragraph{ARC-AGI-1.}
We use the ARC-AGI-1 dataset \citep{chollet2019measure} and follow the HRM and
TRM experimental setup \citep{wang2025hierarchicalreasoningmodel,jolicoeurmartineau2025morerecursivereasoningtiny}.
For RRM, we use $L=2$ and train with
$T_{\mathrm{train}}\in\{8,15\}$ applications and
$N_{\mathrm{sup}}=32$ maximum supervision steps. The $T_{\mathrm{train}}=8$
model uses $T_{\mathrm{detach}}=0$. For $T_{\mathrm{train}}=15$, we train
variants with $T_{\mathrm{detach}}\in\{0,5\}$. At
inference, we evaluate $T_{\mathrm{test}}\in\{8,15,21\}$.
We also evaluate the $T_{\mathrm{train}}=15$, $T_{\mathrm{detach}}=0$ RRM
with $N_{\mathrm{sup}}=2$ at inference instead of the 32 used during training.

\subsection{Learned Policies for Adaptive Ordering}
\label{sec:learned-ordering-appendix}

We also tested replacing uniform layer sampling with a learned routing policy.
At application $t$, the policy samples the next layer from
$\pi_\phi(r_t\mid h_t)$, and the layer parameters are trained with the task
loss. Because the sampled layer indices are discrete, we train the policy with
REINFORCE \citep{williams1992simple}. For a sampled sequence $\mathbf r_i$ and
reward $R_i=-\ell_i$, the policy loss is
\[
  \mathcal L_{\mathrm{policy}}(\phi)
  = -\frac{1}{B}\sum_{i=1}^{B}
  \operatorname{sg}(R_i-\bar R)\sum_{t=0}^{T-1}
  \log \pi_\phi(r_{i,t}\mid h_{i,t}),
\]
where $\bar R$ is the minibatch mean reward.

The policy also has an exit logit, allowing it to stop before the maximum
number of applications. It can therefore learn to reduce computation by using
a smaller $T$. A policy could also concentrate its routes on a subset of the
$L$ layers, making unused layers candidates for removal and potentially
reducing the parameter count.

\begin{wraptable}{r}{0.48\textwidth}
  \vspace{-1.2em}
  \centering
  \small
  \caption{CIFAR-10 learned-ordering ablation. The learned policy's reported
  $T$ is the mean number of applications. $M$ denotes Monte Carlo inference
  samples.}
  \label{tab:cifar10-learned-routing}
  \setlength{\tabcolsep}{3pt}
  \begin{tabular*}{\linewidth}{@{\extracolsep{\fill}}lrrrr@{}}
    \toprule
    Model & $L$ & $T$ & $M$ & Accuracy (\%) \\
    \midrule
    Dense & 8 & $8$ & 1 & $92.52$ \\
    Learned policy & 8 & $7.99$ & 1 & $85.07$ \\
    RRM & 2 & $8$ & 1 & $91.40$ \\
    RRM & 2 & $8$ & 16 & $92.60$ \\
    RRM & 8 & $32$ & 1 & $\mathbf{93.67}$ \\
    \bottomrule
  \end{tabular*}
  \vspace{-0.5em}
\end{wraptable}

However, learning the policy is difficult because it must choose among $L^T$
possible layer orders while the layers themselves are still changing. The
reward is observed only after executing the sampled sequence, which makes
credit assignment across its routing decisions noisy. Table~\ref{tab:cifar10-learned-routing}
shows this behavior in an eight-layer CIFAR-10 ablation. The learned
policy uses an average of $T=7.99$ applications and retains all eight layers,
closely matching the dense model's computation and parameter count while also
adding policy parameters. We also added a penalty on $T$ to encourage shorter
computations, but it substantially degraded accuracy. The learned policy
therefore provides neither a parameter nor a computation benefit over the
dense model and obtains lower accuracy. Uniform random reuse
avoids this additional policy optimization. Learning policies that realize
both computation and parameter savings is left for future work.

\subsection{Inference Procedures}
\label{sec:inference-procedures-appendix}

\subsubsection{Monte Carlo Inference}
\label{sec:mc-inference-appendix}

For a fixed depth $T$, the predictive distribution marginalizes over random
layer sequences. We estimate this expectation using $M$ independent samples,
\begin{equation}
    \begin{aligned}
    p_\theta(y\mid x)
      &= \mathbb{E}_{\mathbf{r}\sim p(\mathbf{r})}
         \left[p_\theta(y\mid x,\mathbf{r})\right], \\
    \widehat{p}_{\theta,M}(y\mid x)
      &= \frac{1}{M}\sum_{m=1}^{M}
         p_\theta\!\left(y\mid x,\mathbf{r}^{(m)}\right),
         \qquad \mathbf{r}^{(m)}\sim p(\mathbf{r}).
    \end{aligned}
    \label{eq:rrm-mc-inference}
\end{equation}
Here, $\widehat{p}_{\theta,M}$ is the $M$-sample Monte Carlo estimator of
$p_\theta(y\mid x)$. Setting $M=1$ uses a single random computation, while
larger $M$ reduces variation across sampled layer sequences at the cost of
additional forward passes, without changing the model parameters.

\subsubsection{PTRM Test-Time Scaling}
\label{sec:ptrm-inference-appendix}

PTRM \citep{sghaier2026probabilistictinyrecursivemodel} produces $K$
stochastic trajectories for the same input. Before each
supervision step $j$ of trajectory $m$, it perturbs the recurrent latent state
with independent Gaussian noise,
\begin{equation}
    \widetilde{h}^{(j,m)}
    = h^{(j,m)} + \epsilon^{(j,m)},
    \qquad
    \epsilon^{(j,m)}\sim\mathcal{N}(0,\sigma^2 I).
\end{equation}
For RRM, each trajectory also uses independently sampled layer sequences. A
trajectory produces a complete candidate solution $\hat{y}^{(m)}$ and a scalar
score $q^{(m)}$ from the existing Q head, which is trained to predict whether
the candidate solution is correct. PTRM returns
\begin{equation}
    m^\star=\arg\max_{m\in\{1,\ldots,K\}}q^{(m)},
    \qquad
    \hat{y}=\hat{y}^{(m^\star)}.
\end{equation}
Unlike Monte Carlo inference, which averages predictive distributions, PTRM
preserves the individual candidates and selects one complete solution. The
approach requires no retraining. We use the main Sudoku-Extreme configuration
of $K=100$, $N_{\mathrm{sup}}=64$ supervision steps, and $\sigma=0.3$. 
Since this method is expensive, we use a fixed subset of 1,000 test puzzles for our evaluations.

\subsection{Additional CIFAR-10 Results}
\label{sec:additional-cifar10}
\begin{figure*}[t]
  \centering
  \begin{minipage}[t]{0.49\textwidth}
    \centering
    \includegraphics[width=\linewidth]{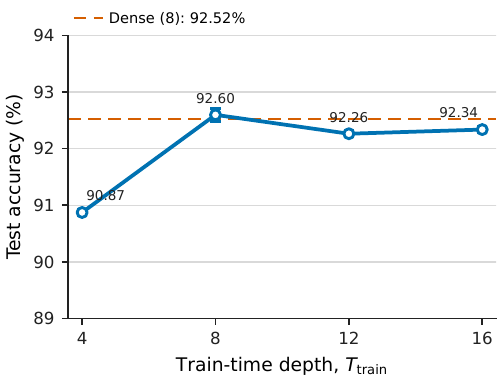}
    \vspace{-0.5ex}
    \small (a) Training-depth scaling
  \end{minipage}\hfill
  \begin{minipage}[t]{0.49\textwidth}
    \centering
    \includegraphics[width=\linewidth]{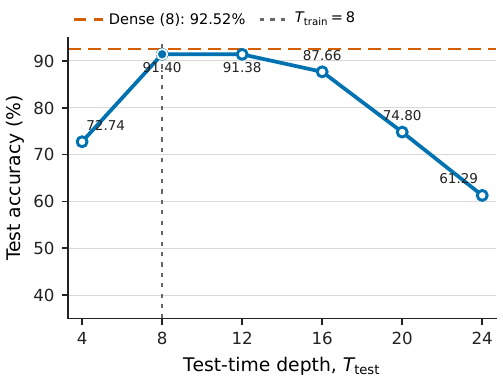}
    \vspace{-0.5ex}
    \small (b) Single-pass test-time scaling, $T_{\mathrm{train}}=8$
  \end{minipage}

  \par\vspace{2.5ex}

  \begin{minipage}[t]{0.49\textwidth}
    \centering
    \includegraphics[width=\linewidth]{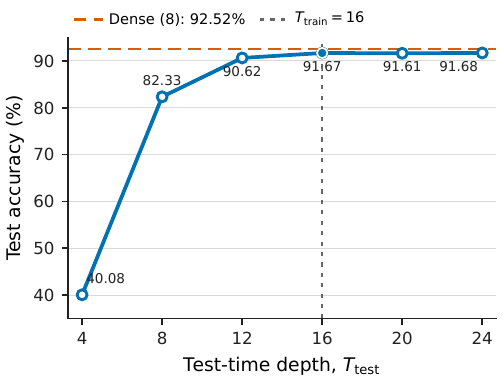}
    \vspace{-0.5ex}
    \small (c) Single-pass test-time scaling, $T_{\mathrm{train}}=16$
  \end{minipage}\hfill
  \begin{minipage}[t]{0.49\textwidth}
    \centering
    \includegraphics[width=\linewidth]{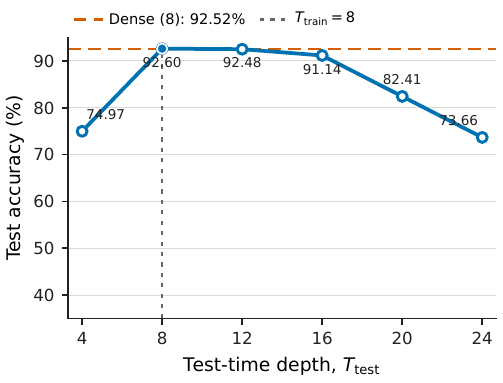}
    \vspace{-0.5ex}
    \small (d) Monte Carlo test-time scaling, $T_{\mathrm{train}}=8$
  \end{minipage}
  \caption{CIFAR-10 training-depth and test-time-scaling experiments. All RRMs
  use $L=2$. Vertical dotted lines mark $T_{\mathrm{train}}$, and
  horizontal dashed lines show the dense eight-layer baseline.}
  \label{fig:cifar10-additional-scaling}
\end{figure*}

\paragraph{Training-depth scaling.}
We train separate $L=2$ RRMs with
$T_{\mathrm{train}}\in\{4,8,12,16\}$ and evaluate each RRM at
$T_{\mathrm{test}}=T_{\mathrm{train}}$ using Monte Carlo inference with $M=16$. Figure~\ref{fig:cifar10-additional-scaling}(a)
shows a clear improvement from four to eight applications, followed by a
plateau. Increasing the training depth alone therefore does not continue to
improve CIFAR-10 accuracy once the model has sufficient depth for the task.

\paragraph{Test-time scaling.}
We next evaluate the RRMs trained with
$T_{\mathrm{train}}\in\{8,16\}$ using one sampled layer sequence per image and
vary $T_{\mathrm{test}}$. Panels (b) and (c) of
Figure~\ref{fig:cifar10-additional-scaling} show that the RRM trained with
$T_{\mathrm{train}}=8$
remains close to its trained-depth performance at $T_{\mathrm{test}}=12$, but
deteriorates when extended further. The RRM trained with
$T_{\mathrm{train}}=16$ is
stable for $T_{\mathrm{test}}\in\{16,20,24\}$. The stable test-time range therefore
depends on the depth encountered during training. Deeper training supports a
wider range of test-time applications, but neither RRM improves
consistently simply by running longer.

\paragraph{Monte Carlo inference.}
Finally, we repeat the test-time sweep for the RRM trained with
$T_{\mathrm{train}}=8$
using $M=16$ sampled sequences per image. As shown in
Figure~\ref{fig:cifar10-additional-scaling}(d), Monte Carlo inference raises the
accuracy at each tested depth, but it does not prevent the degradation far
beyond the training depth. Averaging multiple layer sequences therefore
improves accuracy, but does not make additional test-time applications
consistently beneficial.

These observations should be treated as specific to CIFAR-10. Maze-Hard and
Sudoku-Extreme instead improve beyond their training depths, while Zebra and ARC-AGI-1
largely plateaus. The useful test-time range therefore depends on the task and
training setup, with additional recurrence eventually becoming unhelpful or
detrimental in some settings. This may reflect the \emph{overthinking} behavior
reported in recurrent models
\citep{bansal2022endtoendalgorithmsynthesisrecurrent}.


\subsection{Additional ARC-AGI-1 Results}
\label{sec:additional-arc-agi-1}

\begin{table}[t]
  \caption{ARC-AGI-1 ablations. $T_{\mathrm{detach}}$ is the number of initial applications,
  out of $T_{\mathrm{train}}$, through which gradients are truncated. Total gives
  $T_{\mathrm{test}} \times N_{\mathrm{sup}}$. All models use $L=2$. The TRM-Att result is from
  \citet{jolicoeurmartineau2025morerecursivereasoningtiny}.}
  \label{tab:arc-agi-1-ablations}
  \centering
  \small
  \setlength{\tabcolsep}{3pt}
  \begin{tabular*}{\linewidth}{@{\extracolsep{\fill}}llccccc@{}}
    \toprule
    Model & $T_{\mathrm{detach}}$ & $T_{\mathrm{train}}$ & $T_{\mathrm{test}}$ & $N_{\mathrm{sup}}$ & Total & Pass@2 (\%) \\
    \midrule
    TRM-Att & -- & 30 & 30 & 16 & 480 & $44.60$ \\
    RRM     & 0 &  8 &  8 & 32 & 256 & $42.67 \pm 0.14$ \\
    RRM     & 0 &  8 & 15 & 32 & 480 & $42.79 \pm 0.26$ \\
    RRM     & 5 & 15 & 15 & 32 & 480 & $43.79 \pm 0.29$ \\
    RRM     & 5 & 15 & 21 & 32 & 672 & $43.79 \pm 0.14$ \\
    RRM     & 0 & 15 & 15 &  2 &  30 & $45.79 \pm 0.29$ \\
    RRM     & 0 & 15 & 15 & 32 & 480 & $\mathbf{47.38 \pm 0.25}$ \\
    \bottomrule
  \end{tabular*}
\end{table}

Table~\ref{tab:arc-agi-1-ablations} adds further ARC-AGI-1 ablations.
RRM outperforms the reported TRM baseline at the same total number of applications.
With $N_{\mathrm{sup}}$ reduced from 32 to 2, RRM still outperforms the reported
TRM baseline while using $16\times$ fewer total applications.
We further study gradient truncation during training. In our ARC-AGI-1
ablations, the RRM that
differentiates all applications performs best at the same inference cost.
Increasing $T_{\mathrm{test}}$ from 15 to 21
does not improve the truncated-gradient RRM. Similarly, the model trained with
$T_{\mathrm{train}}=8$ gains only marginally when evaluated at
$T_{\mathrm{test}}=15$.

\end{document}